\documentclass[conference]{IEEEtran}
\IEEEoverridecommandlockouts
\usepackage{hyperref}

\usepackage{cite}
\usepackage{amsmath,amssymb,amsfonts}
\usepackage{algorithmic}
\usepackage{graphicx}
\usepackage{textcomp}
\usepackage{xcolor}
\usepackage{booktabs} 
\usepackage{multirow} 
\usepackage{lipsum} 
\def\BibTeX{{\rm B\kern-.05em{\sc i\kern-.025em b}\kern-.08em
    T\kern-.1667em\lower.7ex\hbox{E}\kern-.125emX}}
\begin{document}

\title{EEG-based Multimodal Representation Learning \\for Emotion Recognition\\

\thanks{This work was supported by the National Research Foundation of Korea (NRF) grant funded by the MSIT (No. 2022-2-00975, MetaSkin: Developing Next-generation Neurohaptic Interface Technology that enables Communication and Control in Metaverse by Skin Touch) and the Institute of Information \& Communications Technology Planning \& Evaluation (IITP) grant, funded by the Ministry of Science and ICT (MSIT) (No. RS-2019-II190079, Artificial Intelligence Graduate School Program).}
}

\makeatletter
\newcommand{\linebreakand}{%
\end{@IEEEauthorhalign}
\hfill\mbox{}\par
\mbox{}\hfill\begin{@IEEEauthorhalign}
}
\makeatother

\author{\IEEEauthorblockN{~~~~~~~Kang Yin}
\IEEEauthorblockA{\textit{~~~~~~~Dept. of Artificial Intelligence} \\
\textit{~~~~~~~Korea University}\\
~~~~~~~Seoul, Republic of Korea \\
~~~~~~~charles\_kang@korea.ac.kr}
\and
\IEEEauthorblockN{~~Hye-Bin Shin}
\IEEEauthorblockA{\textit{~~~~Dept. of Brain and Cognitive Engineering} \\
\textit{Korea University}\\
Seoul, Republic of Korea \\
hb\_shin@korea.ac.kr}
\linebreakand
\IEEEauthorblockN{Dan Li}
\IEEEauthorblockA{\textit{Dept. of Artificial Intelligence} \\
\textit{Korea University}\\
Seoul, Republic of Korea \\
dan\_li@korea.ac.kr}
\and
\IEEEauthorblockN{Seong-Whan Lee}
\IEEEauthorblockA{\textit{Dept. of Artificial Intelligence} \\
\textit{Korea University}\\
Seoul, Republic of Korea \\
sw.lee@korea.ac.kr}
}
\IEEEoverridecommandlockouts
\maketitle

\begin{abstract}
Multimodal learning has been a popular area of research, yet integrating electroencephalogram (EEG) data poses unique challenges due to its inherent variability and limited availability. In this paper, we introduce a novel multimodal framework that accommodates not only conventional modalities such as video, images, and audio, but also incorporates EEG data. Our framework is designed to flexibly handle varying input sizes, while dynamically adjusting attention to account for feature importance across modalities. We evaluate our approach on a recently introduced emotion recognition dataset that combines data from three modalities, making it an ideal testbed for multimodal learning. The experimental results provide a benchmark for the dataset and demonstrate the effectiveness of the proposed framework. This work highlights the potential of integrating EEG into multimodal systems, paving the way for more robust and comprehensive applications in emotion recognition and beyond.
\end{abstract}

\begin{IEEEkeywords}
brain--computer interface, electroencephalogram, multimodal training, emotion recognition;
\end{IEEEkeywords}

\section{INTRODUCTION}
Multimodal representation learning has gained significant attention in the field of artificial intelligence, particularly in tasks that involve complex human behaviors such as emotion recognition~\cite{5}. By combining data from multiple modalities, such as video, audio, and physiological signals, multimodal systems can capture diverse and complementary information to improve model performance~\cite{emo6,8,emo7}. However, integrating electroencephalogram (EEG) data into such frameworks introduces unique challenges due to the inherent variability, noise, and limited availability of EEG datasets compared to other modalities. Despite its potential to provide direct insights into brain activity, the effective utilization of EEG data in multimodal settings remains an open research problem.

EEG data, widely used in neuroscience and clinical research~\cite{1}, offer a non-invasive window into the electrical activity of the brain. This modality has the advantage of capturing cognitive and emotional states in real-time, making it particularly valuable for emotion recognition tasks~\cite{emo4}. However, EEG signals are often noisy~\cite{4}, highly variable across subjects and sessions, and recorded at a higher dimensionality than traditional modalities like video and audio. These characteristics complicate the feature extraction process and hinder the straightforward integration of EEG into multimodal learning frameworks~\cite{emo8}. Existing work on emotion recognition has largely focused on conventional modalities, with many studies opting to exclude EEG due to these challenges~\cite{6}. As a result, the full potential of EEG data in improving emotion recognition systems remains underexplored~\cite{emo2, 9}.
\begin{figure*}[!ht]
    \centering
    \resizebox{0.95\linewidth}{!}{
        \includegraphics{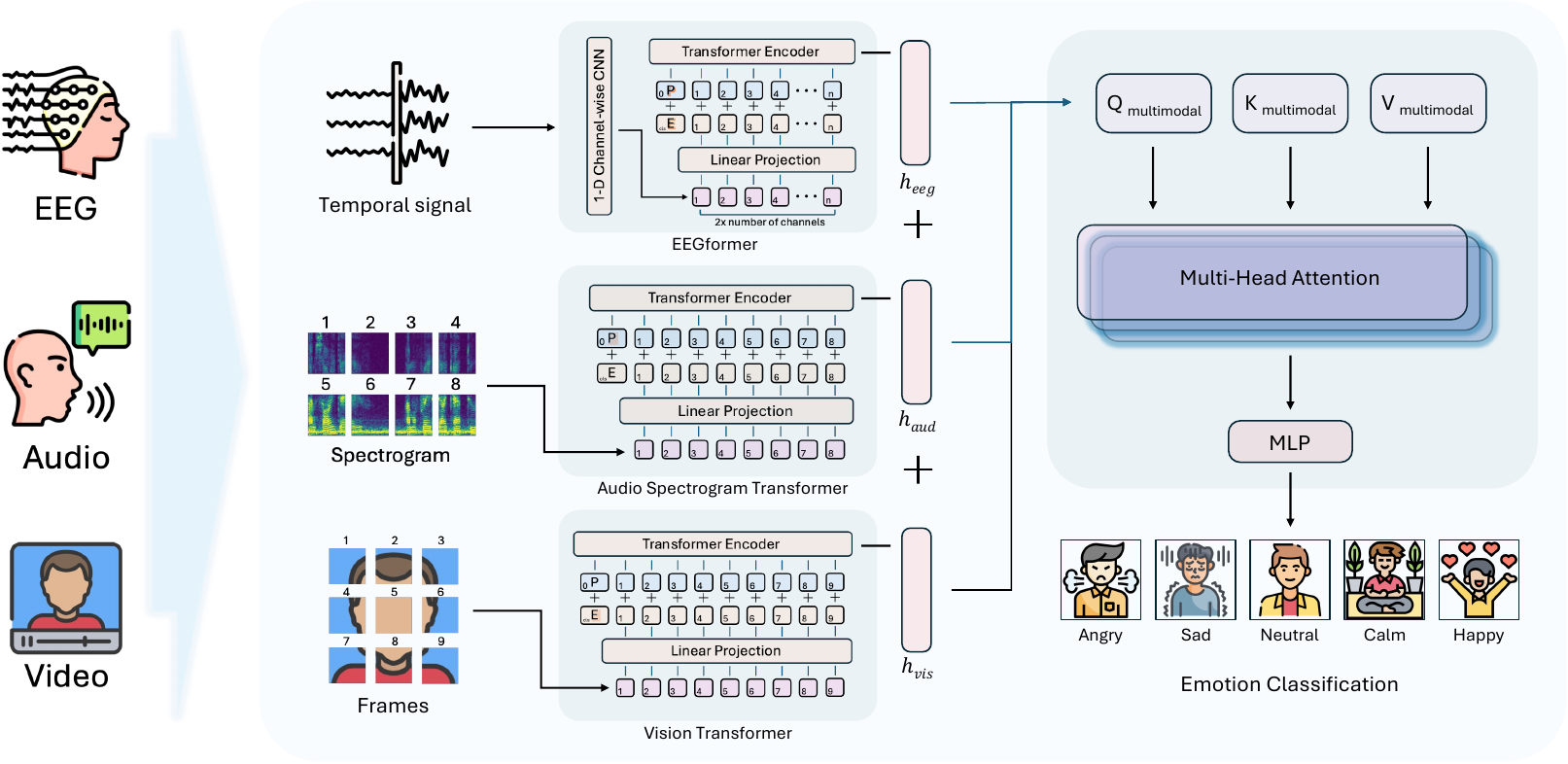}
    }
    \caption{Overview of the proposed attention-based multimodal emotion recognition framework, extracting EEG, audio, and visual features with specifically tailored transformers and integrating them through self-attention fusion.}
    \label{framework}
\end{figure*}
Recent advances in multimodal learning have explored various fusion strategies to integrate heterogeneous data sources, such as feature concatenation~\cite{emo1}, attention mechanisms~\cite{emo5}, and joint embedding spaces~\cite{emo3}. These methods have shown promise in combining modalities like video and audio, where data characteristics are more homogeneous. However, the integration of EEG data presents a distinct set of limitations~\cite{3}. Current models often struggle to dynamically adjust to the variability in feature importance across different modalities~\cite{emo9}, especially when EEG is involved. Additionally, existing multimodal systems may not be flexible enough to handle the varying input sizes and feature distributions that arise when combining EEG with other sensory data~\cite{7}. This often results in suboptimal performance or requires manual tuning to accommodate the unique nature of EEG signals~\cite{emo10, 2}.

In this paper, we propose a novel multimodal framework for emotion recognition that integrates EEG, video, and audio data. Our approach dynamically adjusts attention weights to prioritize key features from each modality and adapts to varying input sizes. Evaluated on a new multimodal emotion recognition dataset, our model sets a new benchmark, highlighting the potential of incorporating EEG in emotion recognition.

By addressing the limitations of existing multimodal frameworks and incorporating EEG data, our work paves the way for more robust and versatile applications in emotion recognition. The proposed framework highlights the value of using diverse data sources to capture complex emotional states, offering a more comprehensive understanding of human emotions than unimodal systems alone. 

\section{METHODOLOGY}

\subsection{Framework Overview}
The proposed multimodal framework, illustrated in Fig.~\ref{framework}, begins by processing video, audio, and EEG data through specifically tailored transformers to extract unique features from each modality. The video input is divided into frames, which are fed into a Vision Transformer (ViT)~\cite{vit} to capture spatial and temporal visual patterns. Similarly, audio is transformed into a spectrogram and passed through an Audio Spectrogram Transformer (AST)~\cite{ast}, designed to extract meaningful frequency and temporal features. EEG data, represented as temporal signals, are processed by the EEGformer~\cite{eegformer}, a transformer-based architecture equipped with a 1-D channel-wise convolutional neural network (CNN) to handle the high-dimensional, channel-specific nature of EEG signals.

Once each modality’s distinct features are extracted—denoted as $h_{vis}$ for video, $h_{aud}$ for audio, and $h_{eeg}$ for EEG—the features are concatenated and sent into a shared multi-head attention module. This attention mechanism allows the model to learn the importance of each modality’s features in a dynamic and context-dependent manner. Through this process, the framework is able to weigh the contributions of visual, auditory, and brainwave data appropriately for emotion recognition. Finally, the fused features are passed through a multi-layer perceptron (MLP) to output the final emotion prediction, classifying emotions such as anger, sadness, neutrality, calmness, and happiness.

\subsection{Network and Implementation}

In the pre-training stage, each modality's transformer is trained separately on its own input domain. This modality-specific training phase allows the network to focus on learning the unique representations inherent to each modality without interference from other data sources. By pre-training the ViT on video frames, the AST on audio spectrograms, and the EEGformer on brainwave signals, we ensure that each feature extractor captures the most salient features relevant to its input type. This modular approach helps mitigate the complexity that arises from combining heterogeneous data sources.

The fine-tuning stage involves fusing the extracted features from all three modalities and processing them together in a joint learning framework. At this stage, we freeze the weights of the pretrained feature extractors to retain their modality-specific learned representations and introduce a shared multi-head attention decoder for further fine-tuning. This attention-based fusion mechanism dynamically adjusts the importance of each modality by attending to the most relevant features based on the context of the emotion recognition task. The modality-specific features $h_{vis}$, $h_{aud}$ and $h_{eeg}$ are first flattened and then input into the multi-head attention decoder, which learns to combine and prioritize these features for optimal emotion classification.

The shared multi-head attention decoder effectively handles the interplay between visual, auditory, and brainwave signals by using query, key, and value representations for each modality, allowing the network to focus on the most informative features for the task. This fusion process ensures that complementary information from the different modalities is leveraged in a synergistic manner, enhancing the overall model's performance in emotion recognition. The output of the multi-head attention layer is a fused multimodal feature representation, which is then passed through a MLP for final emotion classification.

For both the pre-training and fine-tuning stages, we adopt a simple yet effective cross-entropy loss function for classification, given by the following equation:
\begin{equation}
    \label{cls_loss}
    \begin{aligned}
        \mathcal{L}_{cls}=\boldsymbol{y}_h\log f_{}(\boldsymbol{h}),
    \end{aligned}
\end{equation}
where $\boldsymbol{h}$ represents the feature vector (either unimodal or multimodal), $f$ is its corresponding transformer decoder, and $\boldsymbol{y}_h$ is the ground truth class label. This objective ensures that the model learns to minimize the difference between the predicted and actual emotion classes during training. The implementation is available at https://github.com/Kang1121/bci-winter-2025.

\section{EXPERIMENT}
\subsection{Dataset}
The EAV~\cite{eav} dataset, recently released, includes 42 subjects with 30-channel EEG, video, and audio recordings, each contributing 200 interactions, with 20-second trials during both listening and speaking tasks. While 200 EEG trials and video clips are available, only 100 audio files correspond to speaking-only interactions. This is the first public dataset combining EEG, audio, and video for emotion recognition in a conversational setting.

Following the authors' preprocessing methods, we evaluate the performance of classic transformer encoders and our proposed multimodal framework at the subject level.

\subsection{Subject-wise Task Performance}
\begin{table}[ht]
    \centering
    \caption{Comparative results of modalities across subjects (\%).}
    \resizebox{0.9\linewidth}{!}{
        \begin{tabular}{ccccc}
            \toprule
            \textbf{Subject} & \textbf{Vision} & \textbf{Audio} & \textbf{EEG} & \textbf{Multimodal} \\
            \midrule
        1  & 55.20 & 58.33 & 59.17 & \textbf{66.60} \\
        2  & 70.03 & 72.50 & 64.17 & \textbf{76.27} \\
        3  & \textbf{76.43} & 52.50 & 54.17 & 75.43 \\
        4  & 77.43 & 60.00 & 66.67 & \textbf{81.83} \\
        5  & \textbf{62.03} & 50.00 & 40.00 & 59.47 \\
        6  & \textbf{83.83} & 80.00 & 48.33 & 69.73 \\
        7  & 74.77 & 60.00 & 59.17 & \textbf{80.43} \\
        8  & 66.60 & 48.33 & 54.17 & \textbf{68.97} \\
        9  & 62.13 & 65.83 & 45.00 & \textbf{76.43} \\
        10 & 66.43 & 53.33 & 47.50 & \textbf{69.37} \\
        11 & \textbf{59.20} & 45.00 & 44.17 & 57.03 \\
        12 & 51.30 & 48.33 & 45.00 & \textbf{55.17} \\
        13 & 73.43 & 66.67 & 55.00 & \textbf{75.27} \\
        14 & \textbf{58.33} & 52.50 & 33.33 & 57.97 \\
        15 & \textbf{73.80} & 67.50 & 51.67 & 65.53 \\
        16 & 54.97 & 55.00 & 42.50 & \textbf{57.83} \\
        17 & 83.77 & 80.83 & 67.50 & \textbf{89.63} \\
        18 & 67.10 & 56.67 & 64.17 & \textbf{74.97} \\
        19 & 61.50 & 60.00 & 51.67 & \textbf{68.10} \\
        20 & 76.37 & 67.50 & 73.33 & \textbf{86.50} \\
        21 & 71.47 & 50.83 & 60.00 & \textbf{78.10} \\
        22 & 64.57 & 74.17 & 70.83 & \textbf{76.37} \\
        23 & 58.63 & \textbf{66.67} & 50.00 & 64.63 \\
        24 & 70.13 & 67.50 & 76.67 & \textbf{85.13} \\
        25 & 60.50 & 48.33 & 50.00 & \textbf{67.73} \\
        26 & 66.33 & 58.33 & 49.17 & \textbf{68.57} \\
        27 & 82.57 & 54.17 & 65.00 & \textbf{83.87} \\
        28 & 71.97 & 55.83 & 67.50 & \textbf{81.17} \\
        29 & 61.87 & 45.83 & 42.50 & \textbf{62.53} \\
        30 & \textbf{66.70} & 55.00 & 50.00 & 56.10 \\
        31 & 67.73 & \textbf{70.83} & 44.17 & 64.30 \\
        32 & 57.93 & 50.83 & 55.83 & \textbf{63.83} \\
        33 & 76.00 & 76.67 & 62.50 & \textbf{80.10} \\
        34 & 63.60 & 40.83 & 46.67 & \textbf{68.20} \\
        35 & 57.23 & 47.50 & 28.33 & \textbf{62.97} \\
        36 & 62.23 & 60.83 & 60.83 & \textbf{74.23} \\
        37 & 57.20 & 43.33 & 55.83 & \textbf{61.83} \\
        38 & 75.93 & 44.17 & 45.83 & \textbf{76.27} \\
        39 & 67.43 & 62.50 & 50.83 & \textbf{71.50} \\
        40 & 57.07 & 60.00 & 40.00 & \textbf{66.90} \\
        41 & \textbf{78.23} & 53.33 & 43.33 & 72.33 \\
        42 & 73.23 & 55.00 & 65.00 & \textbf{77.10} \\
        \midrule
        Avg. & 67.22 & 58.17 & 53.51 & \textbf{70.86} \\
            \bottomrule
        \end{tabular}
    }
    \label{tab:result}
\end{table}

Table~\ref{tab:result} presents the performance of each unimodal model alongside our proposed multimodal framework, evaluated on a subject-wise basis. We benchmark the EAV dataset using multimodal inputs, achieving a performance of 70.86 \% in accuracy. This represents an improvement of 3.64 \% over the vision-only modality, 12.69 \% over the audio modality, and 17.35 \% over the EEG modality. These results highlight the advantage of integrating multiple modalities, especially given the limitations of unimodal approaches. As previously reported in~\cite{eav}, using unimodal data for emotion recognition—particularly EEG and audio—does not yield highly satisfying results. Our experiments show that EEG data alone achieves an average accuracy of 53.51 \%, while audio data reaches 58.17 \%. In contrast, video data performs significantly better, achieving an accuracy of 67.22 \%, underscoring its prominent role in emotion recognition tasks.

The discrepancy in performance across modalities suggests that visual information carries more distinctive cues for emotional state classification, as video captures a broad range of non-verbal signals such as facial expressions and gestures. From our case studies, we observed that despite the experimental design, which required subjects to remain fully engaged during both speaking and listening tasks, the emotional content in the audio recordings appeared subdued. In contrast, video clips revealed more nuanced differences across emotional classes, such as micro-expressions and body language, which were more reliably captured by the vision-based model. This may explain the superior performance of the video modality relative to audio and EEG data.

Fig.~\ref{fig:result} provides a visual comparison of the subject-wise performance across modalities. The figure illustrates the variability in performance between subjects, reinforcing the idea that emotion recognition is subject-dependent to some extent. Importantly, our proposed multimodal framework consistently outperforms the vision-only transformer across nearly all subjects, demonstrating the effectiveness of our fusion strategy in capturing and integrating the complementary information provided by each modality. By leveraging the strengths of video, audio, and EEG data, the proposed model can make more informed predictions, resulting in overall improved accuracy. These findings confirm that multimodal approaches are crucial in addressing the limitations of unimodal systems and enhancing the robustness of emotion recognition tasks.

\begin{figure}[t]
    \centering
    \resizebox{1\linewidth}{!}{
        \includegraphics{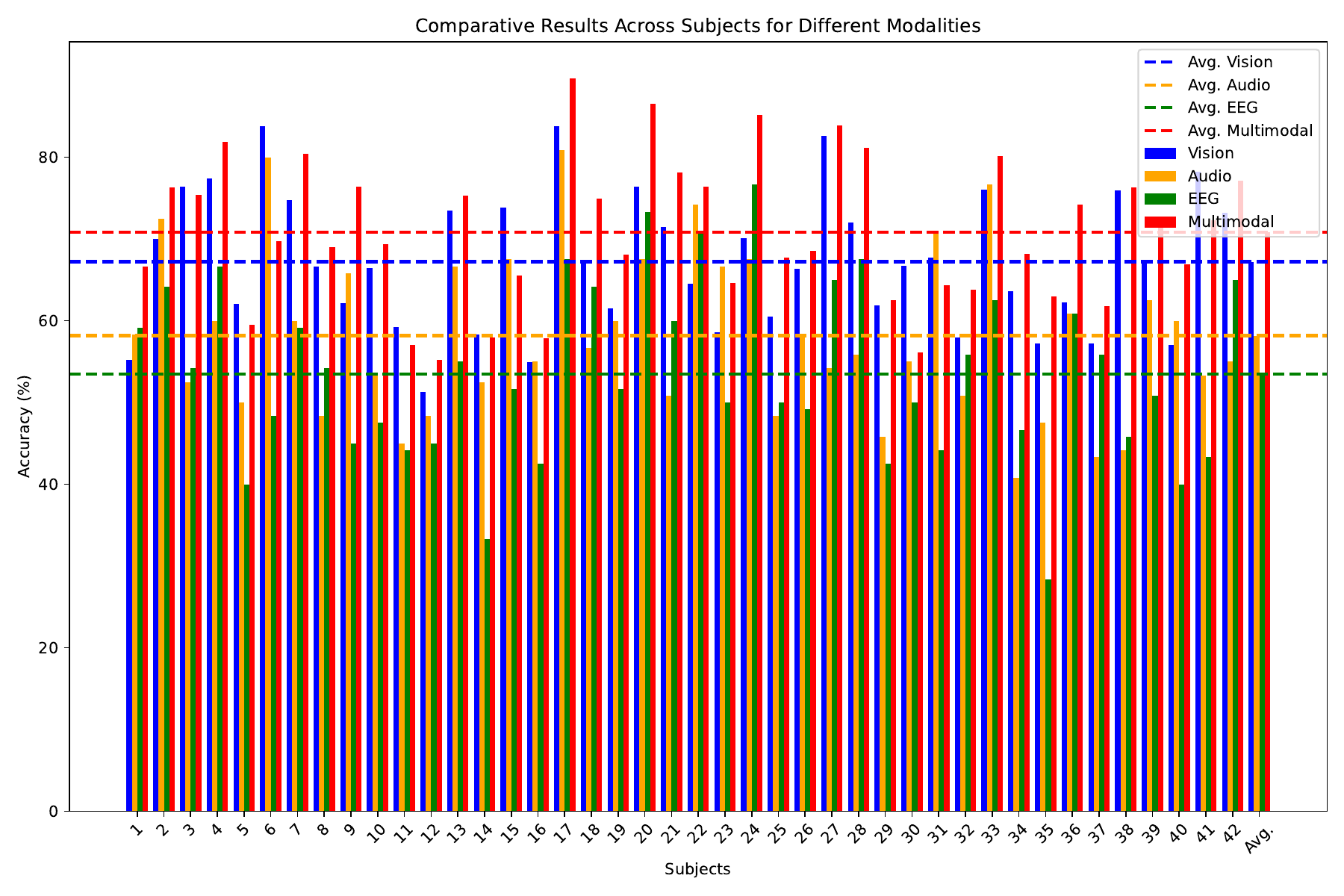}
    }
    \caption{Barplot illustration of the subject-wise performance across modalities.}
    \label{fig:result}
\end{figure}
\section{CONCLUSION}
We propose a two-stage, end-to-end multimodal framework that integrates EEG, video, and audio inputs, enabling the joint learning of different modalities for emotion recognition tasks. Leveraging transformers as the backbone for each modality, our framework effectively captures and fuses modality-specific features, allowing for a more comprehensive understanding of emotional states. The effectiveness of this approach is demonstrated through extensive benchmarking on the newly introduced EAV dataset, which is specifically designed to support multimodal learning for emotion recognition.

Our proposed framework is both simple and highly effective, offering a strong baseline for future researchers to build upon. By presenting this framework, we hope to inspire further exploration into EEG-based multimodal learning, encouraging the research community to not only benchmark on this dataset but also delve deeper into the rich possibilities offered by integrating diverse modalities. Through this work, we aim to contribute a foundational model that can guide future advancements in multimodal emotion recognition and EEG-based research.
\bibliographystyle{jabbrv_IEEEtran}
\bibliography{reference}

\end{document}